\documentclass[]{spie}  

\usepackage{amsmath,amsfonts,amssymb}
\usepackage{graphicx}
\usepackage[colorlinks=true, allcolors=blue]{hyperref}
\usepackage{amssymb}
\usepackage{adjustbox}
\usepackage{graphicx}
\usepackage{siunitx}
\usepackage{mathtools}
\usepackage{threeparttable}
\usepackage{float}
\usepackage{booktabs}
\usepackage{tabularx}
\usepackage{makecell}
\newcolumntype{C}[1]{>{\centering\arraybackslash}p{#1}}
\newcolumntype{L}[1]{>{\raggedright\arraybackslash}p{#1}}
\usepackage{multicol}
\usepackage[T1]{fontenc}
\usepackage{csquotes}
\usepackage{url}
\usepackage{subcaption}

\graphicspath{ 
{fig_money/} 
{fig_results/}
}

\title{3D Universal Lesion Detection and Tagging \\ in CT with Self-Training}

\author[1]{Jared Frazier}
\author[1]{Tejas Sudharshan Mathai}
\author[1]{Jianfei Liu}
\author[2]{\\Angshuman Paul}
\author[1]{Ronald M. Summers}

\affil[1]{Imaging Biomarkers and Computer-Aided Diagnosis Laboratory, Radiology and Imaging Sciences, Clinical Center, National Institutes of Health, Bethesda MD, USA}
\affil[2]{Indian Institute of Technology, Jodhpur, Rajasthan, India}

\authorinfo{Send correspondence to T.S.M.: tejas dot mathai at nih dot gov}

\begin{document} 
\maketitle


\begin{abstract}

Radiologists routinely perform the tedious task of lesion localization, classification, and size measurement in computed tomography (CT) studies. Universal lesion detection and tagging (ULDT) can simultaneously help alleviate the cumbersome nature of lesion measurement and enable tumor burden assessment. Previous ULDT approaches utilize the publicly available DeepLesion dataset, however it does not provide the full volumetric (3D) extent of lesions and also displays a severe class imbalance. In this work, we propose a self-training pipeline to detect 3D lesions and tag them according to the body part they occur in. We used a significantly limited 30\% subset of DeepLesion to train a VFNet model for 2D lesion detection and tagging. Next, the 2D lesion context was expanded into 3D, and the mined 3D lesion proposals were integrated back into the baseline training data in order to retrain the model over multiple rounds. Through the self-training procedure, our VFNet model learned from its own predictions, detected lesions in 3D, and tagged them. Our results indicated that our VFNet model achieved an average sensitivity of 46.9\% at [0.125:8] false positives (FP) with a limited 30\% data subset in comparison to the 46.8\% of an existing approach that used the entire DeepLesion dataset. To our knowledge, we are the first to jointly detect lesions in 3D and tag them according to the body part label.

\end{abstract}





\section{INTRODUCTION}
\label{sec:intro}  

Radiologists routinely size lesions in patient CT studies to estimate tumor burden and stage cancer \cite{Schwartz2016,Gray2015_VariabilityOfLesionSize,Mattikalli2021_EnsembleLesionDetection}. They scroll through the volumetric slices to find the lesion extent, but prospectively measure the lesion size on a single slice to determine malignancy \cite{Gray2015_VariabilityOfLesionSize,Cai2021_LesionHarvester,Yan2021_LENS} according to RECIST guidelines \cite{Schwartz2016}. However, this guideline varies across institutions due to many factors, such as clinical practice, type of CT scanners, exam protocols, contrast phases used. Moreover, the RECIST measurements do not indicate the true 3D lesion extent with regions of the same lesion instance in adjoining slices remaining unmarked \cite{Cai2021_LesionHarvester,Yan2021_LENS,Naga2022_weakSupSelfTraining}. As the 3D lesion extent can be a potential indicator of response to treatment, it is crucial to detect and label the 3D lesions, such that their size can be accurately measured based on current guidelines. Prior approaches \cite{Yan2018_deeplesion,Mattikalli2021_EnsembleLesionDetection,Yang2020_alignShift,Cai2021_LesionHarvester,Yan2021_LENS,Tang2019_uldor,Xie2021_recistnet,Zlocha2019,Yan2018_3dce,Yang2021_A3D,Li2022_SATR} for 3D lesion detection used the publicly available DeepLesion dataset \cite{Yan2018_deeplesion}, but it contains incomplete annotations \cite{Yan2021_LENS,Erickson2022_classImbalanceCorrection,Naga2022_weakSupSelfTraining} and severe class imbalances \cite{Erickson2022_classImbalanceCorrection,Naga2022_weakSupSelfTraining}. 

In this work, we designed a self-training pipeline for 3D lesion detection and tagging. We used a limited 30\% data subset of DeepLesion consisting of lesion bounding boxes and coarse body part labels to train a VFNet model \cite{Zhang2021_vfnet} for 2D lesion detection and tagging. Our model subsequently expanded the context of the 2D boxes + tags in the training set to 3D, and also mapped the 2D boxes + tags predicted on an unseen DeepLesion training set into 3D. These 3D lesion predictions were integrated back into the training data when the confidence scores exceeded a certain threshold to ensure robustness against noise. Self-training was done over multiple rounds, and the proposed model effectively learned from its own predictions and re-trained itself for 3D lesion detection and tagging. Finally, an ensemble of the best performing models across the self-training rounds was used for 3D lesion detection and tagging on a held-out test set. 



\begin{figure}[!t]
\centering
\begin{subfigure}[b]{0.3\columnwidth}
\vspace*{\fill}
  \centering
  \includegraphics[width=\columnwidth,height=3.5cm]{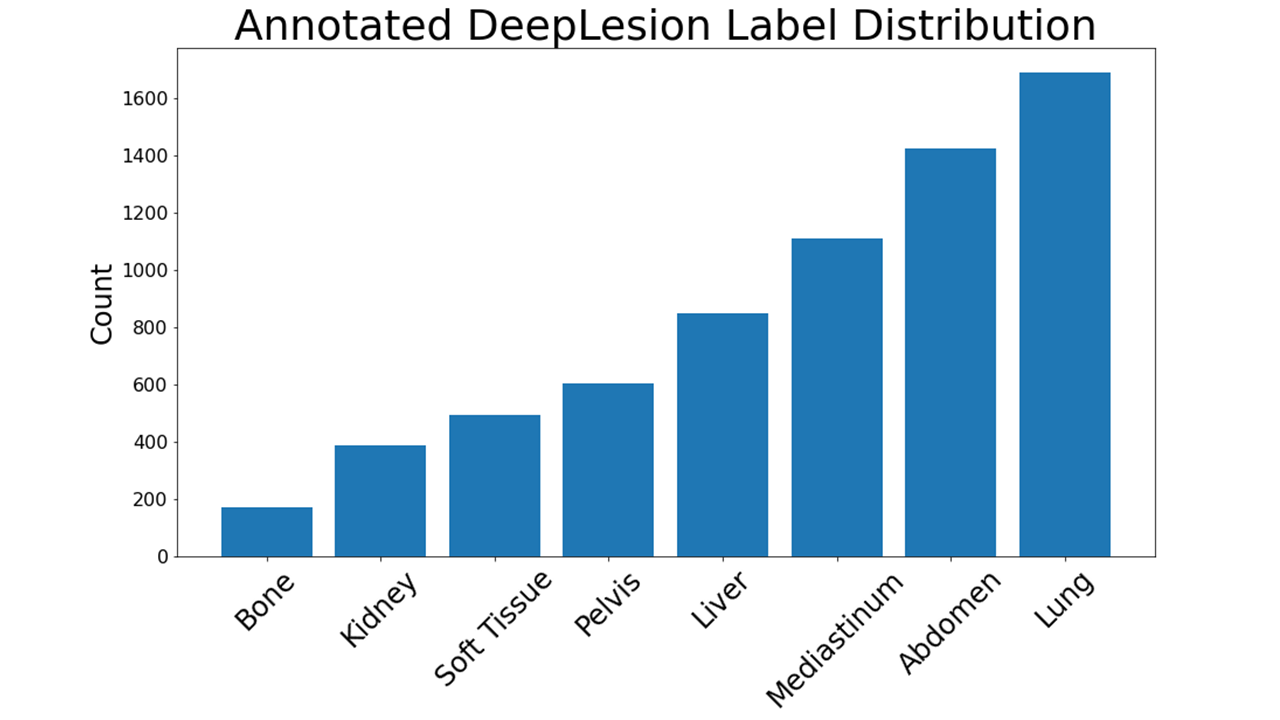}
  \centerline{(a) Label distribution of the DeepLesion} 
\end{subfigure} 
\begin{subfigure}[b]{0.5\columnwidth}
\vspace*{\fill}
  \centering
  \includegraphics[width=\columnwidth,height=3.5cm]{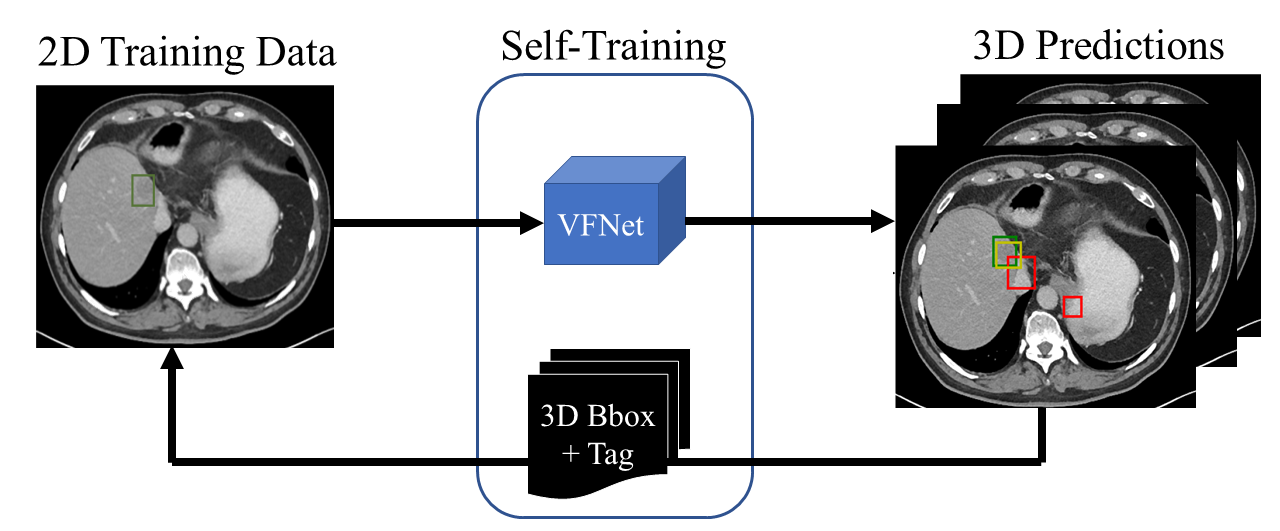}
  \centerline{(b) Self-training pipeline} 
\end{subfigure} 
\begin{subfigure}[b]{0.25\columnwidth}
\vspace*{\fill}
  \centering
  \includegraphics[width=\columnwidth,height=3.5cm]{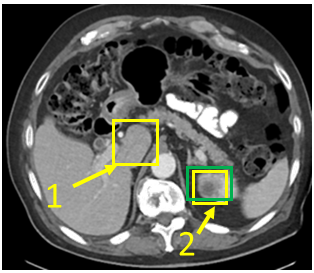}
  \centerline{(c) Slice 53} 
\end{subfigure} 
\begin{subfigure}[b]{0.25\columnwidth}
\vspace*{\fill}
  \centering
  \includegraphics[width=\columnwidth,height=3.5cm]{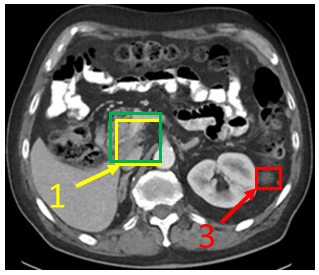}
  \centerline{(d) Slice 59} 
\end{subfigure} 
\begin{subfigure}[b]{0.25\columnwidth}
\vspace*{\fill}
  \centering
  \includegraphics[width=\columnwidth,height=3.5cm]{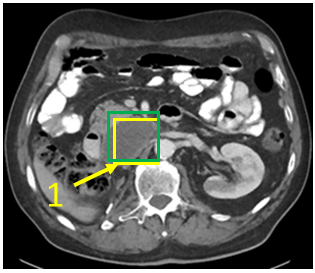}
  \centerline{(e) Slice 64} 
\end{subfigure} 
\caption{(a) Distribution of lesion labels in the DeepLesion validation + test splits showing class imbalances. (b) Proposed self-training pipeline. A VFNet model was trained on 2D lesions annotations from the baseline training split, and it iteratively mined new 2D predictions on the training data and on an unseen data split (original DeepLesion training split). Following this step, the 2D predictions were consolidated based on IoU overlap and expanded into 3D. Next, the 3D predictions were incorporated into the baseline training split for re-training the model over multiple rounds. (c) - (e) 3D predictions with confidence score $\geq$ 50\% shown for different slices of an abdominal CT volume. They were generated by the ensemble of models from different mining rounds self-trained with the variable threshold policy. Green: Ground Truth, Yellow: True Positive (IoU \cite{Yan2021_LENS} $\geq$ 30\%), and Red: False Positive. Lesion 1 is a GT ``abdomen'' lesion spanning across slices $[58, 73]$, while the predicted 3D proposal spans slices $[53, 73]$. Despite GT boxes being absent for lesion 1 in slices 53 through 57, the 3D proposal has an IoU overlap \cite{Yan2021_LENS} $\geq$30\% with the GT and is considered a TP. Notice that lesion 1 is visible across all displayed slices. Lesion 2 is a GT ``kidney'' lesion spanning slices $[52, 57]$ with the predicted 3D lesion spanning slices $[53, 54]$. Lesion 3 is a predicted 3D ``abdomen'' lesion spanning slices $[59, 63]$, and it was counted as a FP during lesion detection + tagging. This lesion was not annotated in the official DeepLesion test split, and thus the tag from this split was not mapped to the fully annotated LENS test split (see Sec.~\ref{sec:methods} for details). Therefore, it is \textit{not} a FP with respect to detection only. Also, the correct tag for lesion 3 is a ``kidney'' cyst, but these lesion types are often confused with the ``liver'' and ``abdomen'' classes. It is not visible in slice 64 due to the predicted extent. 
}
\label{fig:VisualSummary}
\end{figure}
\section{METHODS}
\label{sec:methods}

\textbf{Data.} The DeepLesion dataset \cite{Yan2018_deeplesion} has 10,594 CT studies from 4,427 patients with 32,120 CT slices containing 32,735 lesions annotated with 2D bounding boxes. It was divided into official training (70\%), validation (15\%) and testing (15\%) splits with the lesions in the validation and testing splits tagged with one of 8 coarse body part tags (bone, abdomen, mediastinum, liver, lung, kidney, soft tissue, pelvis). As we aim to achieve ULDT in this work, we use only the validation and testing splits (30\% of total data). The official training split was used for lesion mining and all the annotations were discarded. To generate a 3D test set, we used the LENS \cite{Yan2021_LENS} test split, which contained 1000 sub-volumes randomly drawn from the DeepLesion test split with all lesions in each sub-volume fully annotated with 3D bounding boxes. Note that these 3D bounding boxes were not tagged, and we mapped the tags from the 2D lesion annotations in the official DeepLesion test split to the 3D boxes in LENS test split when the intersection over union (IoU) between them exceeded 10\%. This meant that the 2D annotation in DeepLesion test split corresponded to the same 3D annotation in the LENS test split. After the labels were mapped, any patients in the official test split that overlapped with the 3D LENS test split were removed to prevent data leakage. A summary of the limited dataset we used is shown in Table~\ref{tab:Data}.


\begin{table*}[!htb]
\centering\fontsize{9}{12}\selectfont 
\setlength\aboverulesep{0pt}\setlength\belowrulesep{0pt} 
\setlength{\tabcolsep}{7pt} 
\setcellgapes{3pt}\makegapedcells 
\caption{Summary of label distributions for the limited 30\% DeepLesion dataset used in this study.}
\begin{adjustbox}{max width=\textwidth}
\begin{tabular}{@{} c|c|c|c|c|c|c|c|c|c @{}} 
\toprule
Subset       & Bone & Abdomen & Mediastinum & Liver & Lung & Kidney & Soft tissue & Pelvis & Total \\
\midrule

Train & 97   & 788     & 613         & 426   & 1039 & 195    & 288         & 321    & 3767  \\

Val    & 44   & 318     & 258         & 205   & 345  & 82     & 107         & 189    & 1548  \\

Test  & 31   & 319     & 239         & 217   & 308  & 109    & 99          & 94     & 1416  \\ 

\midrule

Total        & 172  & 1425    & 1110        & 848   & 1692 & 386    & 494         & 604    & 6731  \\ 

\bottomrule
\end{tabular}
\end{adjustbox}
\label{tab:Data}
\end{table*}

\noindent
\textbf{3D Lesion Detection and Tagging.} A recently proposed object detector called Varifocal Net \cite{Zhang2021_vfnet} (VFNet) was used for ULDT. VFNet was trained to predict a lesion's 2D bounding box and body part label in a CT slice. Weighted Boxes Fusion \cite{Solovyev2021_WeightedBoxesFusion} (WBF) was used to consolidate the numerous predictions from multiple epochs of a single model or from multiple models. If the IoU between the predicted 2D boxes across multiple consecutive slices exceeded 30\% \cite{Yan2021_LENS}, they were consolidated together to generate the 3D context for a predicted lesion. In this way, the 3D lesion prediction was made up of a cluster of 2D boxes. The [$x$, $y$] coordinates of the final 3D bounding box were computed as the average of the 2D boxes weighted by their detection scores. The body part tag for the 3D lesion prediction was obtained from the tag of the constituent 2D box with the highest predicted confidence. Implementation details followed that prior work \cite{Mattikalli2021_EnsembleLesionDetection} and details are described in Sec. \ref{suppMat_modelImplementation}.

\noindent
\textbf{Self-Training with Intra- and Inter-Patient Data.} VFNet was used to iteratively mine new 3D lesions in \textit{our} baseline train split (intra-patient) and the \textit{official} DeepLesion training split (inter-patient). Mined 3D lesions that were predicted in these splits were filtered based on exceeding a specified confidence score (see Sec. \ref{sec:results}). These 3D mined lesions were incorporated back into the baseline training split, and the model was re-trained (from scratch) on some of its own predictions. The self-training procedure was repeated for four mining rounds. 

\section{EXPERIMENTS AND RESULTS}
\label{sec:results}


\textbf{Experimental Design.} Two different 3D lesion mining experiments were conducted for incorporating lesion proposals back into the baseline training split. The first experiment $E_{S}$ used a static threshold to include 3D lesion predictions with confidence scores $\geq$ 80\% across the mining rounds. The static threshold was selected under the belief that high quality lesions would be mined that lead to an efficient lesion detector. The second experiment $E_{V}$ varied the confidence score threshold across mining rounds. An initial threshold of 80\% was set for the first round and it was progressively lowered by 10\% across remaining rounds. $E_{V}$ operated under the belief that larger lesion quantities would be mined across rounds to improve detection performance. An additional facet of both experiments was the upsampling of mined 3D proposals in the baseline training set, such that an approximately even distribution of each lesion type was visible to the model during training. First, the number of predicted lesions for the most predominant class in a round was computed, and then the number of lesions of all other classes were upsampled to match that of the prevalent class. The lesion upsampling strategy was adopted because a disproportionate number of mined 3D proposals could further exacerbate the class imbalance in the DeepLesion dataset as seen in Fig.~\ref{fig:VisualSummary}(a). It would also nullify the need to discard mined 3D lesions across over-represented classes in order to balance them against the under-represented classes. We compared these experimental results against those from a model that underwent no self-training. Consistent with prior work \cite{Yan2021_LENS,Cai2021_LesionHarvester}, we present our results of sensitivity at 4 FP/volume, sensitivity at [0.125,0.25,0.5,1,2,4,8] FP/volume, and at 30\% intersection of bounding box (IoBB) overlap \cite{Cai2021_LesionHarvester} on the LENS test split in Table~\ref{tab:PerformanceSummary} and Fig.~\ref{fig:VisualSummary}. 
\begin{table}[!htb]
\centering\fontsize{10}{12}\selectfont 
\setlength\aboverulesep{0pt}\setlength\belowrulesep{0pt} 
\setlength{\tabcolsep}{7pt} 
\setcellgapes{3pt}\makegapedcells 
\caption{3D lesion detection and tagging performance of VFNet. Sensitivites were calculated at 4FP and at 30\% IOBB\cite{Yan2021_LENS} overlap. Averaged sensitivity at [0.125, 0.25, 0.5, 1, 2, 4, 8] FP is also shown.}
\begin{center}
\begin{adjustbox}{max width=\textwidth, max height = 5cm}
\begin{tabular}{@{} c|c|c|c|c|c|c|c|c|c @{}} 
\toprule
Round         & Bone       & Kidney     & Soft Tissue       & Pelvis       & Liver       & Mediastinum       & Abdomen       & Lung      & Sens@0.125:8FP\\
\midrule
\multicolumn{10}{c}{\textbf{No Self-Training (Baseline)}}\\
\midrule
Round 0      & 77.4 & 61.7 & 72.7 & 61.5 & 68.2 & 61.9 & 72.4 & 48.6 & 46.7 \\
\midrule
\makecell{\# Lesions used}  & \makecell{97\\(2.6\%)} & \makecell{195\\(5.2\%)} & \makecell{288\\(7.6\%)} & \makecell{321\\(8.5\%)} & \makecell{426\\(11.3\%)} & \makecell{613\\(16.8\%)} & \makecell{788\\(20.9\%)} & \makecell{1039\\(27.6\%)} & - \\
\midrule
\multicolumn{10}{c}{\textbf{$E_{S}$ -- Static Threshold Policy [80\%, 80\%, 80\%, 80\%] }}\\
\midrule
Round 1 (80\%)        & 64.5 & 56.4 & 76.8 & 54.1 & 66.4 & 56.5 & 69.5 & 48.0 &  44.8 \\
Round 2 (80\%)        & 64.5 & 58.5 & 77.8 & 54.1 & 69.1 & 56.9 & 68.8 & 50.2 & 44.8 \\
Round 3 (80\%)        & 67.7 & 50.0 & 71.7 & 54.1 & 65.4 & 64.0 & 67.9 & 48.6 & 43.1\\
Round 4 (80\%)        & 74.2 & 45.7 & 67.8 & 61.5 & 62.2 & 54.4 & 66.6 & 44.5 & 42.2 \\
Ensemble of Rounds    & 71.0 & 59.6 & 72.7 & 62.4 & 65.9 & 57.7 & 71.1 & 51.4 & 46.5 \\
\midrule
\makecell{\# Lesions mined}  & \makecell{279\\(4.3\%)} & \makecell{245\\(3.7\%)} & \makecell{256\\(3.9\%)} & \makecell{71\\(1.1\%)} & \makecell{756\\(11.5\%)} & \makecell{1084\\(16.5\%)} & \makecell{3347\\(51.0\%)} & \makecell{521\\(8.0\%)} & - \\
\midrule
\multicolumn{10}{c}{\textbf{$E_{V}$ -- Variable Threshold Policy [80\%, 70\%, 60\%, 50\%] }}\\
\midrule
Round 1 (80\%)        & 64.5 & 56.4 & 76.8 & 54.1 & 66.4 & 56.5 & 69.5 & 48.0 & 44.8 \\
Round 2 (70\%)        & 77.4 & 54.3 & 75.8 & 56.0 & 63.6 & 55.2 & 71.1 & 46.7 & 43.9 \\
Round 3 (60\%)        & 64.5 & 50.0 & 73.7 & 57.8 & 65.4 & 56.5 & 65.9 & 42.6 & 42.1 \\
Round 4 (50\%)        & 67.7 & 59.6 & 67.8 & 54.1 & 65.0 & 54.0 & 67.2 & 38.9 & 40.8 \\
Ensemble of Rounds    & \textbf{77.4} & 58.5 & \textbf{75.7} & \textbf{61.5} & \textbf{68.2} & 56.9 & 70.1 & 43.0 & \textbf{46.9} \\
\midrule

\makecell{\# Lesions mined}  
& \makecell{1494\\(2.6\%)} & \makecell{3849\\(6.6\%)} & \makecell{3850\\(6.6\%)} & \makecell{1070\\(1.8\%)} & \makecell{7577\\(12.9\%)} & \makecell{7455\\(12.7\%)} & \makecell{26009\\(44.5\%)} & \makecell{7169\\(12.3\%)} & - \\
\bottomrule
\end{tabular}
\end{adjustbox}
\end{center}
\label{tab:PerformanceSummary} 
\end{table}

\begin{figure}[!htb]
    \centering
    \includegraphics{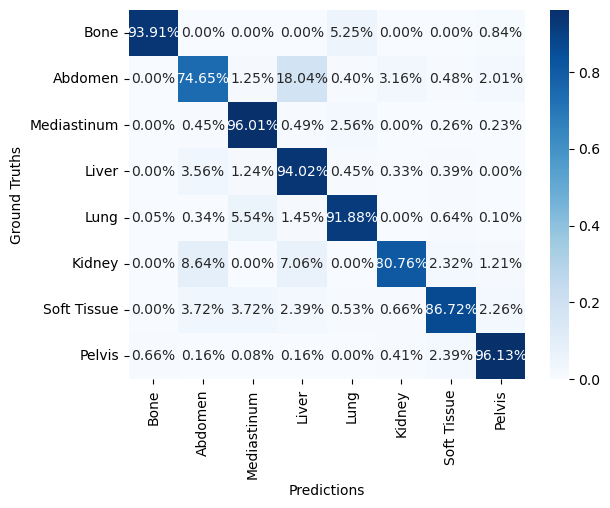}
    \caption{Normalized Confusion Matrix: No Self-Training (Baseline).}
    \label{fig:NoSelfTrainConfusionMatrix}
\end{figure}

\begin{figure}[!htb]
    \centering
    \includegraphics{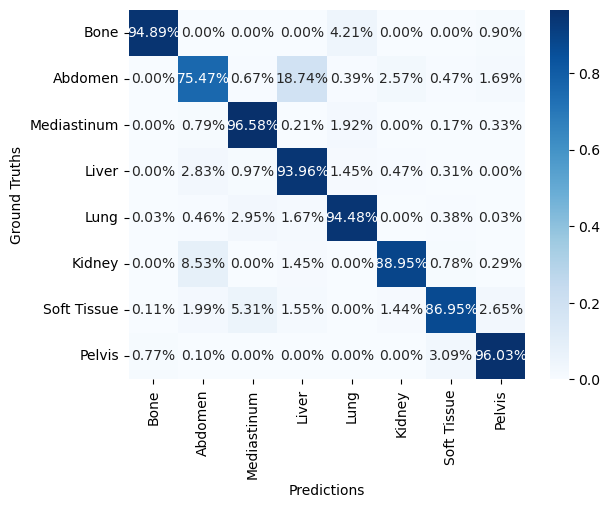}
    \caption{Normalized Confusion Matrix: Ensemble of Static Threshold Models.}
    \label{fig:EnsembleStaticModelsConfusionMatrix}
\end{figure}

\begin{figure}[!htb]
    \centering
    \includegraphics{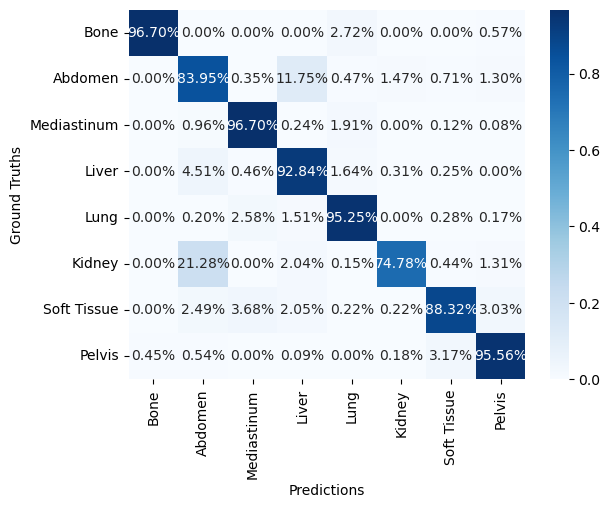}
    \caption{Normalized Confusion Matrix: Ensemble of Variable Threshold Models.}
    \label{fig:EnsembleVariableModelsConfusionmatrix}
\end{figure}

\noindent
\textbf{Results and Discussion.} The VFNet model without self-training achieved a mean sensitivity of 46.7\% at [0.125:8] FP/vol with the highest sensitivities for over-represented classes (abdomen 72.4\%) and the lowest sensitivities seen for under-represented classes (kidney 61.5\%, pelvis 61.7\%). Comparing the static threshold experiment $E_{S}$ against the model with no self-training, the sensitivity at [0.125:8] FP/vol was slightly lower (46.7\% vs 46.5\%). It became evident that only 3/8 classes (kidney, abdomen, and soft tissue) maintained or improved the sensitivity with self-training at 4FP/vol. We believe the ``Abdomen'' class performed poorly because the original DeepLesion dataset \cite{Yan2018_deeplesion} described this class as a ``catch-all'' term for all abdominal lesions that were not ``Kidney'' or ``Liver'' masses. However, these two organs are in close proximity to each other anatomically, and axial slices often show cross-sections of both the liver and kidney in the same slice as seen in Fig.~\ref{fig:VisualSummary}(c). The confusion matrices shown in Figs. \ref{fig:NoSelfTrainConfusionMatrix}, \ref{fig:EnsembleStaticModelsConfusionMatrix}, and \ref{fig:EnsembleVariableModelsConfusionmatrix} confirmed this belief as it showed that the ``abdomen'' lesions were predominantly confused with the ``liver'' masses and ``kidney'' lesions. 

Comparing the variable threshold experiment $E_{V}$ against the model without self-training, we noticed that the average sensitivity at [0.125:8] FP/vol was higher (46.9\% vs 46.7\%). Amongst the under-represented classes (bone, kidney, soft tissue, pelvis), 3/4 classes maintained or improved their performance. Only the ``kidney'' class showed a $\sim$3\% drop in sensitivity at 4FP. Among the over-represented classes, we observed that only 1/4 classes (liver) maintained or improved their detection performance with self-training. The remaining classes saw a small decrease in sensitivity at 4FP. The model from $E_{V}$ performed similarly when compared against the results described in prior work \cite{Yan2021_LENS} on the same test dataset (46.9\% vs 46.8\%). Specifically, we show similar performance with a \textit{significantly} smaller dataset (30\% of original DeepLesion) in contrast to prior work \cite{Yan2021_LENS} who used the entire Deeplesion dataset in conjunction with models trained on additional datasets\cite{Setio2017_LUNA, Bilic2019_LITS, Roth2014_NIHLN} for lesion mining. 

\section{CONCLUSIONS}
\label{sec:conclusions}

In this work, we developed a self-training pipeline for 3D universal lesion detection and tagging. We used only a 30\% fraction of the DeepLesion dataset to train a VFNet model for 2D ULDT. Following this, the model simultaneously expanded the 2D boxes and tags into 3D for the training split and an unseen data split (official DeepLesion training split). The 3D proposals were then incorporated into the baseline training data when the confidence scores of the proposals exceeded a confidence threshold. Self-training ensued with the model learning from its own predictions over multiple rounds. The final trained model was used for evaluation, and predicted 3D lesions and their tags for a separate test split. We have shown that our average sensitivity of 46.9\% at [0.125:8] FP/vol is comparable to 46.8\% of prior work \cite{Yan2021_LENS} where the entire DeepLesion dataset was used.


\acknowledgments 

This work was supported by the Intramural Research Program of the NIH Clinical Center.

\bibliography{main_references} 
\bibliographystyle{spiebib} 

\newpage
\section{SUPPLEMENTARY MATERIAL}
\label{sec:supplementary}

\subsection{Implementation}
\label{suppMat_modelImplementation}

The window center and width (e.g. [-175,275]) values provided in the DeepLesion dataset were used to window the Hounsfield units (HU)
in a CT slice and clip them to the [0,255] range. Additionally, slices above and
below the annotated slice were also windowed and normalized. To mimic
the radiologist’s approach of scrolling through slices in a CT volume, we constructed a 2.5D image with three consecutive slices (middle slice was the annotated slice) for training the detectors. Each slice was resized to 512×512 pixels
dimension before being fed to the detectors. The backbone network for VFNet was a ResNet-50 model, and data was augmented through
standard strategies, such as random flipping, rotation, intensity shifts, crops. The batch size was set to 2, the learning rate was $1e^{-3}$, and VFNet was trained for 16 epochs. Experiments were run on a workstation running Ubuntu 18.04 LTS with 4 Tesla V100 GPUs. Fusion of predictions through WBF was done at test time with the top 5 epochs having the lowest validation for each model type with the exception of the ensemble of self-trained models. For the ensemble of self-trained models, the epoch having the lowest validation loss for each self-trained model was used for fusion of 2D predictions. Consistent with literature, models were evaluated at 0.30 IOBB threshold\cite{Yan2021_LENS}. The IOBB threshold is computed by calculating the intersection of predicted bounding boxes and ground truth bounding boxes and then dividing by the area of the predicted bounding boxes (as implemented in the \href{https://github.com/viggin/DeepLesion_manual_test_set}{github} of prior work\cite{Yan2021_LENS}).


\end{document}